\documentclass[a4paper, graybox]{svmult}

\usepackage{type1cm}        
\usepackage{makeidx}         
\usepackage{graphicx}        
\usepackage{multicol}        
\usepackage[bottom]{footmisc}

\usepackage{newtxtext}       %
\usepackage[varvw]{newtxmath}       

\usepackage{subfigure}
\usepackage{url}

\usepackage{array}
\usepackage{hyperref}

\usepackage{eso-pic}
\newcommand\AtPageUpperMyright[1]{\AtPageUpperLeft{%
 \put(\LenToUnit{0.5\paperwidth},\LenToUnit{-1.5cm}){%
     \parbox{0.5\textwidth}{\raggedright\fontsize{11}{11}\selectfont #1}}}}
\newcommand{\conf}[1]{\AddToShipoutPictureBG*{\AtPageUpperMyright{#1}}}

\makeindex             

\conf{\hspace*{-5cm}\textcolor{gray}{This paper has been accepted for publication at \textcolor{blue}{\href{https://iser2023.org/instructions-for-authors/}{(ISER 2023)}}. }}

\begin{document}
\title*{POVNav: A Pareto-Optimal Mapless Visual Navigator}
\titlerunning{POVNav} 
\author{Durgakant Pushp$^{1}$, Zheng Chen$^{1}$,  Chaomin Luo$^{2}$, Jason M. Gregory$^{3}$ and Lantao Liu$^{1}$}
\institute{
 $^{1}$D. Pushp, Z. Chen and L. Liu are with Luddy School of Informatics, Computing, and Engineering, Indiana University, Bloomington, IN 47408, USA. Email: {\tt\small dpushp@iu.edu} \newline
 $^{2}$C. Luo is with the Department of Electrical and Computer Engineering, Mississippi State University, MS 39762, USA. 
 \newline
 $^{3}$J. Gregory is with DEVCOM Army Research Laboratory, USA. 
}
%
%

\maketitle

\vspace{-3.cm}
\section{Introduction}
\label{sec:1}
\vspace{-0.3cm}

Mapless visual navigation has introduced a promising alternative for autonomous navigation, eliminating the reliance on building explicit maps~\cite{chaplot2020neural, gupta2017unifying, Ort2018MNFramework}. Despite the significant advancements in visual detections~\cite{Strudel_2021_ICCV, CALI_2022_RSS, chen2023} that enable the identification of navigable areas (e.g., road, grass, mulch, etc.) and non-navigable areas (e.g., tree, car, house, etc.), their application in local motion planning tasks remains relatively unexplored.
In this work, we propose a vision-based local planning and control framework, a Pareto-optimal mapless visual navigator (POVNav), 
which utilizes segmented navigability perspective images as an immediate environmental representation for real-time navigability analysis. 
The local planning and control module, based on a Pareto-optimal strategy, selects a sub-goal in the image, plans a safe visual path, and generates appropriate actions using visual servo control. Our approach enables selective navigation behavior, such as restricting navigation to selected terrain types, by modifying the navigability definition in the local representation. The ability of POVNav to navigate a robot to the goal using only a monocular camera makes it computationally light and easy to implement on various robotic platforms.
Our contributions include the development of the navigability image, which enhances the navigational flexibility by incorporating semantic meaning. This representation relies on the concept of the visual horizon to guide the planning process. However, this novel representation also poses challenges related to defining the decision space, embedding goal information on the image, formulating the objective function in the image space, and generating efficient motion from image features. We address these challenges in the POVNav framework.
Real-world experiments in diverse challenging environments, ranging from structured indoor environments to unstructured outdoor environments such as forest trails and roads after a heavy snowfall, using various image segmentation techniques demonstrate the remarkable efficacy of our proposed framework. Furthermore, the comparative analysis between POVNav and existing image-based planner~\cite{chen2023polyline} highlighted the superior performance of our proposed framework in terms of run-time efficiency, path length, and success rate across various simulated environments.
We provide open-sourced code for further evaluation: \url{https://github.com/Dpushp/POVNav}. 
\vspace{-0.7cm}
\section{Related Work}
\label{sec:related_work}
\vspace{-0.3cm}
Our work is closely related to mapless navigation, which is broadly comprised of non-learning (optical flow, appearance-based, object feature tracking), and learning-based solutions. We provide a brief review of them as follows.

\textbf{Geometry and feature-based approaches:}
Prior research has explored optical flow features \cite{ OF_ttt}, stored image templates \cite{AB_Bista}, information-theoretic approach \cite{AB_Dame} 
and feature tracking-based navigation methods~\cite{FT_Anad2, Srivastava_2021, Pushp_2022} to control robot motion. 
An image segmentation method based on sub-region growing to find paths is presented in \cite{SB_Zhang} while control parameters are obtained using path boundary information. 
These methods address the problem of obstacle avoidance but are inadequate to achieve goal-oriented navigation. 
Vision-based and goal-oriented navigation is challenging as the possible features that can be extracted from the image fail to directly be interpreted in terms of the goal point, which is essentially a 3-D point in the space. 
{\em The proposed work not only finds the optimal sub-goal in the local observation but also correlates the visual features with the goal point.} Moreover, different from existing methods, we aim to design a unified framework that allows a robot to navigate in an environment with no a-priori information by only relying on a monocular camera and an approximate goal pose.

\textbf{Learning based approaches:} 
A large family of current frameworks is data-driven or based on machine learning
\cite{shen2019situational,bansal2020combining}.
For example, imitation learning-based approaches have been broadly explored to train a navigation policy that enables a robot to mimic human behaviors or navigate closely to certain waypoints without a {prior} map~\cite{manderson2020vision,hirose2020probabilistic}. 
A large amount of work on visual navigation can also be found in the computer vision community~\cite{shen2019situational, chaplot2020neural, gupta2017unifying}. 
These approaches exploit and integrate deep learning methods to train navigation policies that perform remarkably well when provided sufficient training data, but can fail if no or very limited data is available.
In contrast to the data-driven approaches, we introduce a computationally light but effective planning and control loop that decouples the representation from the behavior learning (like modular learning approaches) using segmentation-based visual representation. Our solution is more flexible, robust, and adaptable to a variety of navigation missions since it makes no assumption about the data it needs to learn the policy. 

\vspace{-0.7cm}
\section{Technical Approach}
\label{sec:technical_approach}
\vspace{-0.3cm}

\begin{figure*}[t!]
    \centering
    \subfigure[POVNav]
        {\label{fig:framework} \includegraphics[height=1.5in, width=3.35in]{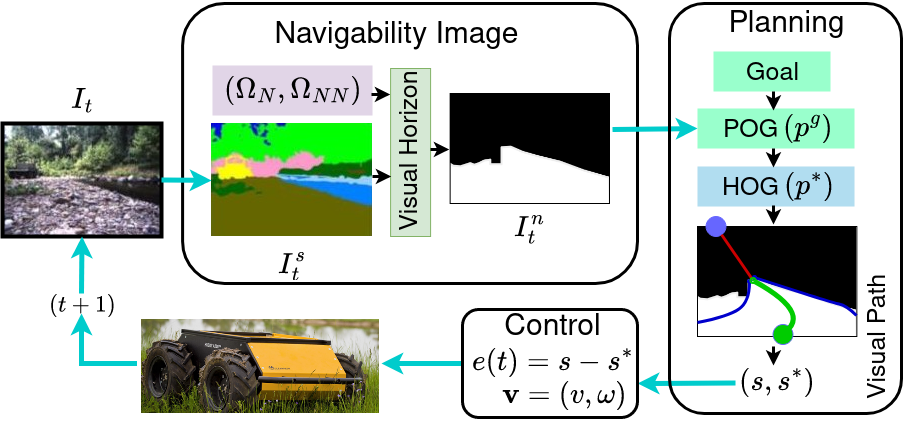}}
    \subfigure[]
        {\label{fig:planning_space} \includegraphics[height=1.5in, width=1.1in]{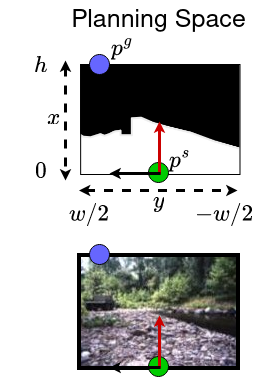}}
    \caption{\small Illustration of our proposed framework, POVNav. (a) The information flow from the input image ($I_t$) to control signals ($v, \omega$) is depicted. $I^s_t$ represents the segmented image, $\Omega_N$ and $\Omega_{NN}$ are navigable and non-navigable class labels. $I_t^n$ is the navigability image constructed by implementing a {\em Visual Horizon} (see Sec. \ref{sec:visual_rep}). The planning module maps the given goal to the border of $I_t^n$ as POG ({\em Peripheral Optic Goal}). HOG ({\em Horizon Optic Goal}) represents a pareto-optimal sub-goal in the navigable space, selected through a multi-objective optimization process over image pixels. The features ($s, s^*$) generated by visual path planning are utilized to determine the control signals. 
    (b) The planning image (top) and its corresponding input RGB image (bottom) are shown. $p^s$ is the robot's current state (approximated). The coordinate frame's origin is located at $p^s$, with the red arrow representing the x-axis and the black arrow representing the y-axis.\vspace{-0.4cm}}
    \label{fig:decision_space}
\end{figure*}

Image pixels are the 2-D projections of 3-D world, and therefore any pixel may be a potential location that a robot moves to. 
The proposed framework, POVNav, consists of two main components: a {\em Navigability Image} and a {\em Visual Planner}. 
The Navigability Image offers a simple and efficient representation of the intricate real-world environment. The Visual Planner addresses the competing navigation objectives by initially identifying a {Pareto-optimal} visual sub-goal. The subsequent utilization of visual path planning facilitates the generation of navigation features that drives the robot through visual servoing, as depicted in Fig.~\ref{fig:framework}( \href{https://www.youtube.com/watch?v=_lJjTj0Xd3k}{see video explanation}).
We describe these critical modules in this section.

\textbf{Planning Space and Coordinate Frame:} As shown in Fig.~\ref{fig:planning_space}, any point $p$ in the planning space whose $x$ and $y$ coordinates are given as $p_x$ and $p_y$, respectively, is always considered with respect to a fixed reference frame $\mathbf{B}$ whose origin is at $p^s$. Note that our planning space consists of a pixel whose $x,y$ coordinate in the image is a 2-D point. Hence, we use point or pixel interchangeably.

\vspace{-0.7cm}
\subsection{Navigability Image - Binary Visual Local Representation}
\label{sec:visual_rep}
\vspace{-0.3cm}
\begin{figure}[t!]
    \centering
    \includegraphics[width=6.4cm]{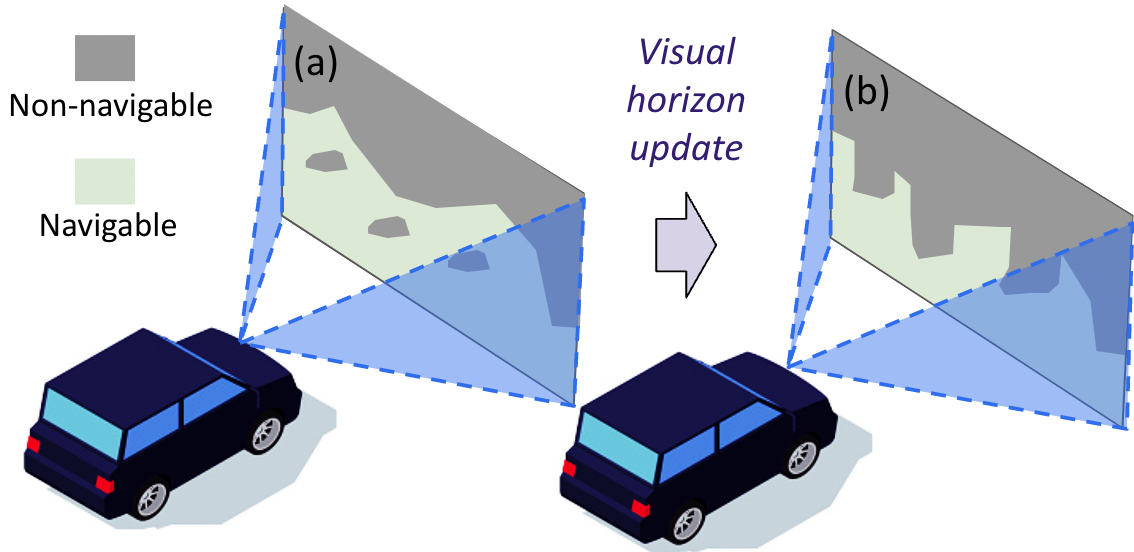}
    \caption{\small
    A perspective binary segmentation image can include a scenario: non-navigable regions surrounded by navigable ones.
    \vspace{-0.4cm}}
    \label{fig:horizon_update}
\end{figure}

Let $I_t^s$ be the segmented image obtained using any off-the-shelf segmentation method $\Gamma$ such that $(I_t^s, \Omega) = \Gamma(I_t)$ where $\Omega := (\Omega_N, \Omega_{NN})$ is the vector of classes defined by $\Gamma$ to classify the pixels. 
We divide $\Omega$ into two groups - navigable classes $\Omega_N$, e.g., trail, grass, mulch, asphalt, and non-navigable classes $\Omega_{NN}$, e.g., sky, tree, bush, building. Using $\Omega$, we define $I_t^b \subseteq \left\{ 0, 1 \right\}^{w\times h}$, where $0$ represents the navigable space, $1$ represents the non-navigable space, and $w, h$ are the width and height of the image, respectively. 

Typically, we obtain some non-navigable segments inside the navigable segments as shown in Fig.~\ref{fig:horizon_update}(a).  We assume that anything behind the non-navigable segments is occluded (or obstructed) and hence considered as non-navigable (see Fig.~\ref{fig:horizon_update}(b)). We further assume all the ``hole-like" navigable surface that is surrounded by non-navigable segments are also non-navigable as they are simply out of reach from the robot's current location. These assumptions are required to minimize the boundary pixels in $I_t^s$ as the boundary pixels have conflicting class labels. The assumptions we propose are reasonable because the regions behind the obstacles will be updated to be navigable or non-navigable as the robot approaches to those regions. To simplify this process, we introduce the concept of {\em Visual Horizon}.  

\textbf{Visual Horizon:} The Visual Horizon $\mathcal{H}_t \in \mathbb{R}^{n \times 2}$, where $n$ is the number of boundary pixels, is defined as a set of contiguous pixels in $I_t$ such that for each $x \in [-w/2, w/2]$ there exists one and only one point $p (x, y)$ for which all the points above are non-navigable and all points below are navigable,
\begin{multline}
\label{Eq:horizon}
~~~~~~~~~~~~~~~~~~~~~~~~~~~~~~~~~~~~~~~~\mathcal{H}_t \coloneqq  \{p(\psi_{t}(y), y)\},\,\,\,\,\,\,\,\,\, \forall \,\,\, y \in [-w/2, w/2], \\ 
    \text{where }\psi_t(y) = x, \,\,\, \text{iff }\,\,\, x \in [0,h), \ s.t. 
    \begin{cases}
        I^b_t(x-i, y) = 0, \,\, \\ 
        ~~~~~~\forall \,\, i \in (1, x] \\
        I^b_t(x+i, y) = 1,  \,\, \\ 
        ~~\forall \,\, i \in (1, h-x].
    \end{cases}
\end{multline}
We obtain navigability image $I^n_t$ by applying Eq.~\eqref{Eq:horizon} on $I^b_t$ that combines all the navigable and reachable segments together as $\mathcal{I}^{N}$ and, all the disconnected navigable segments along with non-navigable segments as $\mathcal{I}^{NN}$. Note that the line $\mathcal{H}_t$ partitions $I_t^n$ into two subsets of pixels - $\mathcal{I}^{N}$ and $\mathcal{I}^{NN}$. 

\vspace{-0.7cm}
\subsection{Pareto-Optimal Sub-goal Selection}
\label{sect:subgoal_selection}
\vspace{-0.3cm}
Given $I^n_t$ and a goal pose with respect to the robot's body frame, we find $p^* \in \mathcal{I}^N$ as the immediate sub-goal to achieve the navigation task when the goal lies outside of the camera's field of view. Thus, we discuss the decision making process to find an optimal sub-goal in the image space. 

There are two basic requirements for local motion planning - the awareness of the robot current state $p^s$ and the goal state $p^g$ in the planning space. As we move from 3-D world to 2-D decision space (i.e., from real world to image), we need to find image pixels in $I^n_t$ that represent these planning variables. Finding the robot's current state is impossible from a first-person-view (FPV) image unless any part of the robot's body is visible in the image. Hence, we find $p^s$ by minimizing the distance between the robot's current state in 3-D and the corresponding 3-D transformations of the image pixels.
This transformation requires depth information to convert 2-D pixels into 3-D points. It is obtained upon camera installation and calibration. For a fixed camera setup, the robot's approximated state $p^s$ in the image  remains constant. In our experiments, it is at the bottom-center pixel according to the camera mount position as shown in Fig.~\ref{fig:planning_space}. 

Next, to find $p^g$ on the image, we map the goal direction from 3-D space to the image border.
This mapping process can be achieved in various ways, and we impose no constraints on the specific mapping technique employed. In our implementation, we put all the values ranging from $[-\pi/2, \pi/2]$ on the left, right, and upper borders of the image and remaining values on the lower border of the image. 
We opted to use border pixels of $I_t^n$ as a representation of $p^g$, as they are the farthest observable pixels in the image, hence they form the boundary of the camera's field of view. 
Therefore, we call $p^g$ a {\em Peripheral Optic Goal} (POG).
It is noteworthy that other pixels in $I_t^n$ could also be chosen to represent $p^g$, as no specific constraint exists regarding the selection of the pixel.

The sub-goal selection involves a multi-objective decision-making process where the decision space is comprised of image pixels. Finding befitting objectives for navigation tasks is a practically challenging issue. In unobstructed environments, optimizing over only one objective, such as shortest path (minimum travel cost), can often suffice. However, in cluttered environments, we need to perform optimization over at least two objectives due to the decision dilemma, e.g., moving toward the goal but also keeping certain clearance from obstacles. We define two objectives over the image space to solve the visual navigation task as

\textbf{Objective 1:} 
Objective $f_1 : \mathcal{I}^N\rightarrow\mathbb{R}$ is defined as the {\em deviation} of a navigation point $p \in \mathcal{I}^N$ from $p^g$ given as:
\begin{equation}
    f_1(p)=|tan^{-1}(p_y / p_x) - tan^{-1}(p^g_y /p^g_x)|.
\end{equation} 

\textbf{Objective 2:} Objective $f_2 : \mathcal{I}^N\rightarrow\mathbb{R}$ is defined as the {\em distance} traveled by the robot in the free space from its current position. As there exists a positive correlation between the distance of two pixels in the image plane and the distance of their 3-D transformations, $f_2$ is thus defined as the Euclidean distance between $p$ and $p^s$ given by:
\begin{equation}
    \begin{aligned}
        f_2(p) = [{(p_x - p^s_x)^2 + (p_y - p^s_y)^2}]^{1/2}.
    \end{aligned}
\end{equation}

Given the two objectives, we define the sub-goal selection task as minimizing $f_1$ and maximizing $f_2$ given by:
\begin{equation} \label{eq:mop_problem}
\begin{aligned}
p^* = 
    \begin{cases}
        &\min\,\,\{f_1(p)\} \, \, \\
        &\max\,\,\{f_2(p)\} \, \, \\
    \end{cases}
    &\textrm{s.t.} \,\, 
    \begin{split}
        -w/2 \leq p_y \leq w/2, \,\, \\
        \forall\,\, p_x \in [0, \psi_t(p_y)], 
    \end{split}
\end{aligned}
\end{equation}
where $-w/2$ and $w/2$ are the lower bound and the upper bound on the decision variable $p_y$, respectively.
These two objectives are often conflicting in nature when the robot encounters obstacles in the scene. 

To address the competing decisions, we base our solution on Pareto optimization. The key concept is based on \textit{dominance} between multiple optimal solutions. 
Formally, a solution $A$ is better than, or dominating, $B$, denoted by $A \succ B$ (or $B \prec A$), if and only if $A$ is not worse than $B$ in all the objectives and, $A$ is strictly better than $B$ in at least one objective.
The non-dominated set of the entire feasible decision space is called the Pareto-optimal set. The boundary defined by the set of all points mapped from the Pareto optimal set is called the {Pareto-optimal front}.

\begin{remark} 
{Given $(f_1, f_2)$ defined over the feasible set $\mathcal{I}^N$, the Pareto optimal solutions must lie on the visual horizon $\mathcal{H}_t$. 
\label{remark1}
}
\end{remark}

This can be easily proven if we consider each pixel on the visual horizon $\mathcal{H}_t$ and compare it with all pixels vertically below it. Each pixel on $\mathcal{H}_t$ dominates  all other pixels in the same column. Therefore, Pareto-optimal pixels must lie on $\mathcal{H}_t$. Readers are encouraged to see the video explanation: \url{https://rb.gy/ib1e8}. 

Remark~\ref{remark1} restricts the search space to $\mathcal{H}_t$. We further narrow the Pareto set $\mathcal{X}^* \subseteq \mathcal{H}_t$ by evaluating the dominance of each element of $\mathcal{H}_t$. 
The problem of picking a solution from the Pareto front for robot navigation should be addressed separately. In this work, we select the optimal sub-goal using the following minimization of the weighted sum of both objectives as
\begin{equation}
\label{Eq:patero_min}
\begin{aligned}
p^* = &\arg\min_{p}  \left(
    w_1 f_1(p) + w_2 \Bar{f_2}(p)
    \right)  \\
&\textrm{s.t. } p \in \mathcal{P};\  \mathcal{P} \subseteq \mathcal{H}_t, \\
\end{aligned}
\end{equation}
where $p$ is the point taken from the Pareto optimal set $\mathcal{P}$ and $\Bar{f_2} = \lVert p^g - p \rVert$ is used for converting $f_2$ into minimization problem, and $w_1$, $w_2$ are the weights given to $f_1$ and $\Bar{f_2}$, respectively. 
We can view $p*$ as the optimal sub-goal in $\mathcal{I}^N$, and we call this sub-goal a {\em Horizon Optic Goal (HOG)} as it lies on the visual horizon defined in the image space. 

\vspace{-0.7cm}
\subsection{Dynamic Feature Generation}
\begin{figure}[t!]
    \centering
        \includegraphics[width=5.8cm, height=3.4cm]{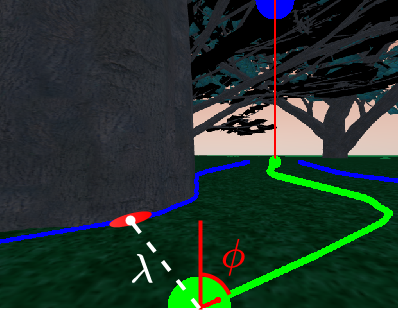}
  	\caption{\small An illustration of a visual path (green line) planned from the start position (green disc) to the Horizon Optic Goal (blue disc) with a safe boundary (blue lines) that accounts for the size of the robot and a safety margin. The red disk is the closest point (obstacle) and the dashed line shows its distance from the robot in the image space. \vspace{-10pt}}
    \label{fig:feature_gen}
\end{figure}

\vspace{-0.3cm}
We formulate goal-oriented navigation as a visual servoing problem that aims to minimize an error $e(t)$, which is defined as $e(t) = s(I^n_t, p^*) - s^*$, where function $s$ provides the features at time $t$, and $s^*$ is the desired features. Note that the HOG which essentially is a point defined over image space, may be considered as a visual feature to control the steering of robot. However, directly following the HOG can lead to a collision with obstacles. Therefore we dynamically generate a path in the image from an approximated start location to the HOG ($p^*$) by calculating a collision-free region based on the robot's footprint, as depicted in Fig.~\ref{fig:feature_gen}. Due to the reduced search space inherent to the image, any path planning algorithm can be employed to effectively determine a path.

We select two features from the dynamically generated visual path - {\em proximity feature} and {\em alignment feature}. The
proximity feature $\lambda$ that captures the relative distance of the closest pixels on the visual horizon, is defined as 
$\text{min}\{\|p - p^s\| : p \in \mathcal{H}_t\}$. 
The alignment feature $\phi$ is defined as the deviation of the visual safe path from the ideal path (see Fig.~\ref{fig:feature_gen}). Basically, $s$ is a function that takes $I^n_t$ and $p^*$ as input and provides $\lambda$ and $\phi$. Therefore, $e(t)$ is modified as $e(t) = [\lambda, \phi] - [\lambda_0, 0]$, where $\lambda_0$ represents the safe proximity and the desired alignment is $0$.  


The error 
is then used to control the robot with a visual servo control scheme \cite{VS_Hutchinson} to generate the camera's 
motion. 
Specifically, we consider a generic robotic system that can be controlled by linear and angular velocities $(v, \omega)$. Let $(v_c, \omega_c)$ be the camera velocity. We use proximity features to find $v_c$ and alignment features to find $\omega_c$. If the camera is fixed on the robot, $v$ and $\omega$ can be obtained from $(v_c, \omega_c)$. 

\vspace{-0.7cm}
\section{Simulation Results}
\label{results}
\vspace{-0.3cm}
In this section, we evaluate POVNav's performance in challenging simulated scenarios. Experiments are conducted in cluttered environments involving scenarios demanding optimization across multiple objectives. Evaluation metrics include success rate, path length, and computation time.

\textbf{Testing Environments: }We consider five unique testing environments, labeled as Env1 to Env5, with increasing difficulty levels. The environments are characterized by varying obstacle proximities, starting with a separation of $3$ meters in Env1 and reducing it to $1$ meter in Env5.

\textbf{Baseline: }We compare our method with the Dynamic Window Approach (DWA)~\cite{fox1997dynamic} due to its wide adoption in reactive planning. DWA confines the search space to collision-free circular trajectories that can be executed within a short time frame. It relies on evaluating numerous trajectories using predefined objective functions to assess planning performance. To ensure a fair comparison, we employ our Navigability Image as the local representation for both methods. Depth information is used in DWA to map candidate trajectories onto the images, as it requires knowledge of obstacle proximity. We refer to this approach as the image-DWA (i-DWA) planner~\cite{chen2023polyline}.

\vspace{-0.7cm}
\subsection{Evaluation of planning performance}
\vspace{-0.3cm}
\begin{figure*}[t!]
     \centering
     \subfigure[SR vs Environments]{
         \includegraphics[height=1.in, width=1.455in]{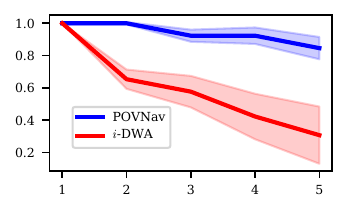}
         \label{fig:C2}}
     \subfigure[PL vs Navigation tasks]{
         \includegraphics[height=1.in, width=1.455in]{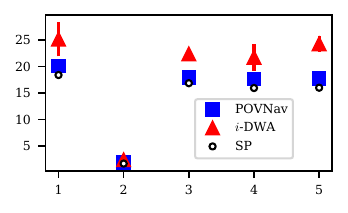}
         \label{fig:C3}}
     \subfigure[PL vs Navigation tasks]{
         \includegraphics[height=1.in, width=1.455in]{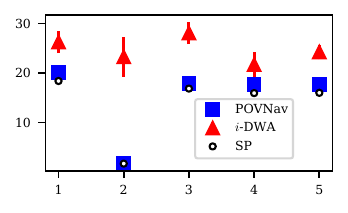}
         \label{fig:C4}}
    \caption{\small Illustration of the performance analysis of the proposed method. (a) X-axis represents different testing environments and Y-axis represents success rate. It presents the comparison results of the proposed POVNav method and the comparative i-DWA method across various environments. In (b) and (c), X-axis represents path length and Y-axis represents different navigation tasks. These subfigures demonstrate the path length comparison of the two methods with respect to different navigation tasks.     
    \vspace{-0.4cm}
    }
    \label{fig:compute_sucess}
\end{figure*}

\begin{figure*}[t!]
  \centering
  \subfigure[Sparse environment (Env3)]{
    \label{fig:env1}
    \includegraphics[height=1.in, width=1.485in]{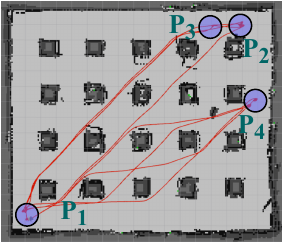}} 
  \subfigure[Dense environment (Env5)]{
    \label{fig:env2}
    \includegraphics[height=1.in, width=1.485in]{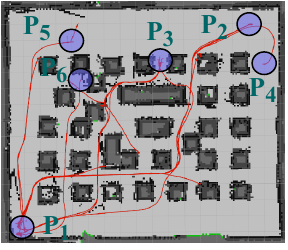}}
  \subfigure[Run-time Comparison]{
    \label{fig:C1}
    \includegraphics[height=1.in, width=1.485in]{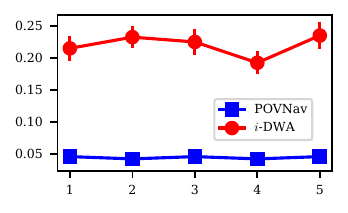}}
  \caption{\small (a) and (b) illustrate the maps of the simulated environment used for validation. Red lines are the path followed by POVNav while performing different navigation tasks. Y-axis in (c) represents time (in seconds) to compute action ($v, \omega$) from observation ($I_t$) during planning, and X-axis represents different environments. 
  \vspace{-10pt}}
\label{fig:three graphs}  
\end{figure*}
We conduct an empirical evaluation of the success rate of the baseline and the proposed method across all the testing environments. We select $20$ randomly generated start and goal points in each environment and use both methods to navigate the robot. The results are presented in Fig.~\ref{fig:C2}, which indicates that the proposed method outperforms the baseline in all the environments, and the performance gap between the two methods gradually widens as the environments become more cluttered. One plausible explanation for this observation is that the baseline methods rely on the depth information to validate the trajectories, and therefore can only perceive obstacles up to the maximum range of the depth sensor, whereas our method plans on the visual horizon, which allows it to take preventive action for collision avoidance even before the baseline is aware of the presence of the obstacles.

In addition to evaluating success rate, we also consider path length. We selected five paths in Env3 for evaluation, denoted as $(\mathbf{p_1}\mathbf{p_2})$, $(\mathbf{p_2}\mathbf{p_3})$, $(\mathbf{p_3}\mathbf{p_1})$, $(\mathbf{p_1}\mathbf{p_4})$ and $(\mathbf{p_4}\mathbf{p_1})$, out of which $(\mathbf{p_2}\mathbf{p_3})$ is obstacle-free. The paths include the largest distance goal, obstacle-free goal, and intermediate point goals. These paths are referred to as navigation tasks 1 to 5. As the environment is homogeneous, these points are sufficient for the comparison studies. An additional set of 5 paths from a dense environment (Env5) namely, $(\mathbf{p_1}\mathbf{p_2})$, $(\mathbf{p_3}\mathbf{p_1})$, $(\mathbf{p_1}\mathbf{p_4})$, $(\mathbf{p_6}\mathbf{p_1})$ and $(\mathbf{p_5}\mathbf{p_1})$ are selected.
We perform $10$ trials for each path. Our comparative analysis reveals that POVNav demonstrates a shorter path than i-DWA, except in cases where both methods have taken the shortest path due to the lack of obstacles. The efficacy of i-DWA is highly reliant on the spacing between the trajectories, as it necessitates converting the continuous velocity space of the dynamic window to discrete space and tuning the control parameters for optimal performance. For densely cluttered environments, a large number of trajectories need to be evaluated. Notably, i-DWA's performance does not generalize well to different environments, as the control parameters optimized for Env1 did not produce similar outcomes in other environments. In contrast, POVNav shows less sensitivity to environmental changes, implying that it can be utilized in various environments without significant parameter tuning.

After evaluating the path lengths, we performed an additional evaluation of the performance of POVNav and the baseline method with respect to computational time, as presented in Fig.~\ref{fig:C1}. The results indicate that POVNav outperforms the baseline method in terms of run-time efficiency, with an average computation time of $0.04$ seconds per action, compared to $0.22$ seconds for the baseline. The efficiency gain can be attributed to the fact that POVNav actively plans the path in the image space and avoids the need to evaluate multiple candidate trajectories in 3-D space as in the case of the baseline. This advantage makes POVNav more suitable for real-time robotic applications where computational efficiency is critical.


To better analyze the trade-off of a navigability-based representation compared to the traditional geometric-based representation, we also compare the proposed Navigability Image with commonly used Occupancy grids.
We found that both representations facilitate effective static and dynamic obstacle avoidance. However, the Navigability Image representation enables selective navigation behavior through perception-aware-planning, which is a critical component for advanced navigation in complex environments.
In addition, the Navigability Image representation is advantageous in terms of relying only on a simple monocular camera, 
making it suitable for low-cost robots. 
Note that, adding semantic meaning to the occupancy grid representation is a possible solution to the issue of ambiguous obstacle detection. However, this approach is susceptible to the depth noise present in real-world environments. Furthermore, it incurs additional computation overhead as it requires projecting 2-D pixels to 3-D space, and it necessitates the use of multiple sensors (such as LiDAR and camera), rendering it unsuitable for resource-constrained robots.

\vspace{-0.7cm}
\subsection{Importance of POG and HOG in POVNav}
\vspace{-0.3cm}

\begin{figure}[t!]\vspace{-8pt}
  \centering
  \subfigure[Obstacle-Free Navigation]
  	{\label{fig:A1}\includegraphics[height=0.93in, width=1.485in]{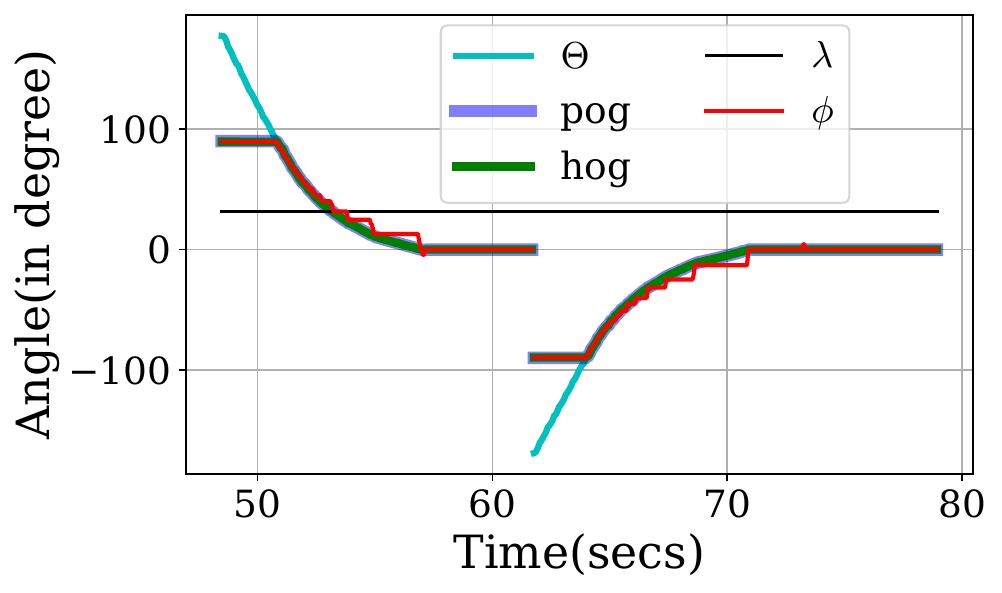}}
  \subfigure[Global Guidance]
  	{\label{fig:Tracking_a}\includegraphics[height=1.in, width=1.485in]{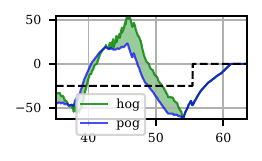}}
  \subfigure[Local Avoidance]
  	{\label{fig:Tracking_b}\includegraphics[height=1.in, width=1.485in]{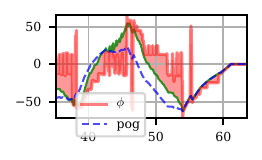}}
  \subfigure[Angular Velocity Control]
  	{\label{fig:Tracking_c}\includegraphics[height=1.in, width=1.485in]{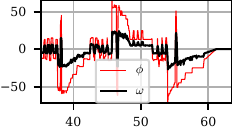}}
  \subfigure[Linear Velocity Control]
  	{\label{fig:Tracking_d}\includegraphics[height=1.in, width=1.485in]{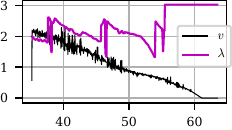}}
  \caption{\small Performance of different components in POVNav. The X-axis in all the subplots represents time and Y-axis represents error value in (b) to (e). (a) shows obstacle-free goal navigation with POVNav ($\lambda = \lambda_0$) for two goals ($\pi$, $-\pi$). (b) Dashed line shows the obstacle detection status (non-zero value means obstacle detected). (c) Shows the effect of visual path planning on obstacle avoidance. (d) Shows $\phi$ controls $\omega$. (e) Shows the effect of $\lambda$ on $v$.
  \vspace{-0.2cm}
  }
\label{fig:tracking}  
\end{figure}

The POG maps the 3-D goal direction to a 2-D image border and guides the robot to the goal analogous to how a navigator uses a compass for global guidance. We validate this claim by performing an experiment in which we define different goals in random directions of an obstacle-free, simulated environment. 
As shown in Fig.~\ref{fig:A1}, the robot aligns itself towards the goal direction and then it maintains the heading angle until it reaches the goal. Note that $\Theta, pog, hog, \phi$ overlaps except for two intervals ($\pi~to~\pi/2, -\pi~to~-\pi/2$) as a result of the non-uniform mapping of the goal direction on the image border. This shows that the implemented mapping function maintains the correlation of POG with $\Theta$. In our case, the POG updates itself with the same value during that interval. This also shows that the POG itself is able to navigate the robot in an obstacle-free environment and can be considered a global guide in POVNav.

We add obstacles in the environment to see how POG, HOG and Visual Planning work together to drive the robot to the goal while avoiding the obstacles. Fig.~\ref{fig:Tracking_a} shows that the sub-goal selection effectively balances the trade-off (represented by the green region) between two conflicting objectives ($f_1, f_2$) and guides the robot to the goal. Note that the alignment of HOG with POG shows maximizing $f_1$. After selecting a sub-goal, the robot dynamically interacts with the environment with visual path planning to actively avoid the obstacles as shown in Fig.~\ref{fig:tracking}.

\vspace{-0.7cm}
\section{Real-World Experiments}
\vspace{-0.3cm}
\begin{figure*}[t!] 
    \centering
    \includegraphics[width=4.6in, height=1.3in]{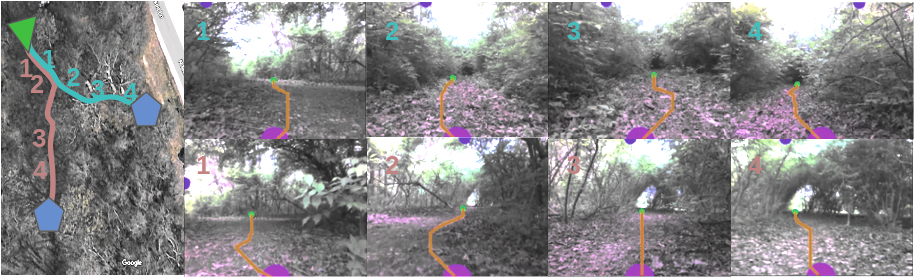}
    \caption{\small Navigation using POVNav in unstructured environments. A green marker indicates the robot's start position. The blue markers denote the goal positions. The purple marker in the images represents the approximated robot state ($s_0$). The orange lines are the planned safe paths. \vspace{-0.4cm}}
    \label{Fig:rw_exp_forest_trail1}
\end{figure*}

\begin{figure*}[t!]
     \centering
     \subfigure[Road $=$ Navigable; Snow $\neq$ Navigable]{
         \includegraphics[height=1.in, width=1.455in]{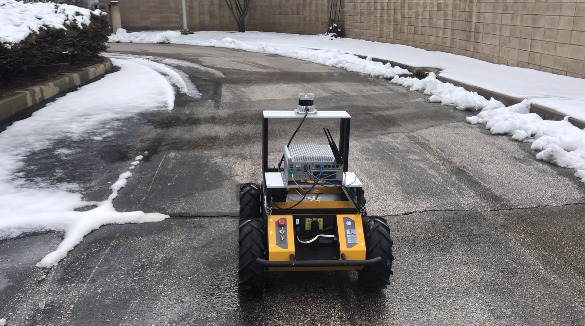}
         \label{fig:out_exp1}}
     \subfigure[Road $\neq$ Navigable; Snow $=$ Navigable]{
         \includegraphics[height=1.in, width=1.455in]{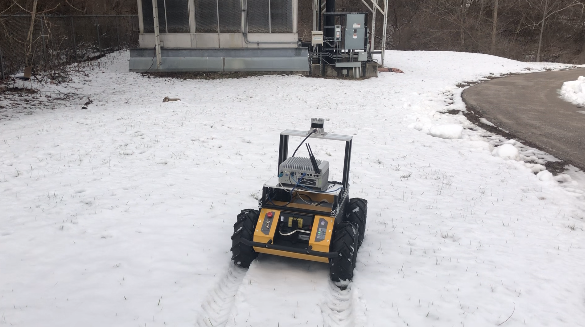}
         \label{fig:out_exp2}}
     \subfigure[Road $=$ Navigable; Snow $=$ Navigable]{
         \includegraphics[height=1.in, width=1.455in]{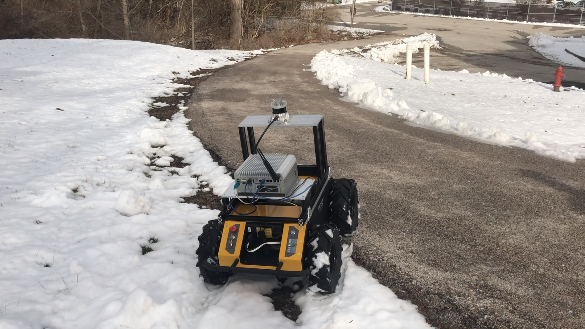}
         \label{fig:out_exp3}}
    \caption{\small Demonstration of selective navigable behaviors of POVNav. Snow is set to be non-navigable and road is considered as navigable in (a). Snow is set to be navigable and road is considered as non-navigable in (b). Both snow and road are considered as navigable in (c).  
    \vspace{-0.4cm}
    }
    \label{fig:Sel_nav}
\end{figure*}
To evaluate the advantages of POVNav, we validate our approach in difficult real-world indoor and field environments under different environmental conditions (fall and winter) using various image segmentation techniques.

\textbf{Experimental Setup: }We implemented the POVNav framework on a Clearpath's Jackal robot for indoor experiments and on a Clearpath's Husky robot for outdoor experiments. Both robots utilize RGB images obtained from an Intel Realsense D435i camera and state estimation from an Intel Realsense t265 camera. We use surface normal-based segmentation\footnote{\url{https://github.com/Dpushp/depth2surface_normals_seg}} for indoor experiments and an unsupervised learning-based method~\cite{CALI_2022_RSS} for outdoor experiments. 

\textbf{Indoor Experiments:} 
Despite using noisy segmentation, the robot successfully reaches the goal most of the time due to POVNav's active path planning.

\textbf{Outdoor Experiments: }
We select two trails in a dense forest as shown in Fig.~ \ref{Fig:rw_exp_forest_trail1} and define a goal at $(x = 20 m, y = 10 m)$. Fig. \ref{Fig:rw_exp_forest_trail1} shows the snapshots of the path planned by the robot at five different waypoints. We have repeated the experiment $5$ times on both trails. The robot achieves the goal every time on trail 1 (left trail). However, it failed to achieve the goal 2 times on trail 2. This is because the segmentation accuracy was insufficient on trail 2 which is more challenging. 
Note that improving the accuracy of segmentation is not in the scope of this paper; we believe our method can be improved accordingly given improved segmentation solutions.

\textbf{Selective Navigation Behavior: }
The modular architecture of POVNav enables the modification of the navigability definition vector $\Omega$ that is utilized to generate the Navigability Image. This capability facilitates the demonstration of selective navigation behavior by altering $\Omega$. Fig.~\ref{fig:Sel_nav} depicts three scenarios in which the robot navigates using distinct $\Omega$ definitions. At present, human knowledge is required to provide these definitions in POVNav (experimental video: \url{https://t.ly/_JNtb})).

\vspace{-0.7cm}
\section{Conclusion} 
\label{sec:conclusion}
\vspace{-0.3cm}
In this work, we propose an efficient and versatile visual navigation framework, POVNav, for local collision-free navigation in both structured indoor and unstructured outdoor environments using a monocular camera. We design a multi-objective optimization to enable autonomous navigation to the specified goal while avoiding obstacles. We have extensively evaluated the proposed framework on two robotics platforms in simulation and in real-world experiments, including challenging forest trails. The results demonstrate the high efficiency and effectiveness of POVNav in real unstructured environments. Moreover, our framework provides a convenient tool for computer vision researchers to extend their work in the robotics domain. Future work includes incorporating learning the navigability definitions vector through real-world interactions and extending the visual horizon definition to facilitate perception-aware planning by retaining the Pareto-optimality property.

\vspace{-\baselineskip} 

{\small
\bibliographystyle{plain}
\bibliography{ref}

\begin{thebibliography}{10}

\bibitem{FT_Anad2}
V.~Anand, D.~Pushp, R.~Raj, and K.~Das.
\newblock Gaussian mixture model (gmm) based object detection and tracking using dynamic patch estimation.
\newblock In {\em 2019 IEEE/RSJ International Conference on Intelligent Robots and Systems (IROS)}, pages 4474--4481, 2019.

\bibitem{bansal2020combining}
S.~Bansal, V.~Tolani, S.~Gupta, J.~Malik, and C.~Tomlin.
\newblock Combining optimal control and learning for visual navigation in novel environments.
\newblock In {\em Conference on Robot Learning}, pages 420--429. PMLR, 2020.

\bibitem{AB_Bista}
S.~R. Bista, P.~R. Giordano, and F.~Chaumette.
\newblock Appearance-based indoor navigation by ibvs using line segments.
\newblock {\em IEEE Robotics and Automation Letters}, 1(1):423--430, 2016.

\bibitem{OF_ttt}
C.~Boretti, P.~Bich, Y.~Zhang, and J.~Baillieul.
\newblock Visual navigation using sparse optical flow and time-to-transit.
\newblock In {\em 2022 International Conference on Robotics and Automation (ICRA)}, pages 9397--9403, 2022.

\bibitem{chaplot2020neural}
D.~S. Chaplot, R.~Salakhutdinov, A.~Gupta, and S.~Gupta.
\newblock Neural topological slam for visual navigation.
\newblock In {\em IEEE/CVF Conference on Computer Vision and Pattern Recognition}, pages 12875--12884, 2020.

\bibitem{VS_Hutchinson}
F.~Chaumette and S.~Hutchinson.
\newblock Visual servo control. i. basic approaches.
\newblock {\em IEEE Robotics \& Automation Magazine}, 13(4):82--90, 2006.

\bibitem{chen2023polyline}
Z.~Chen, Z.~Ding, D.~J. Crandall, and L.~Liu.
\newblock Polyline generative navigable space segmentation for autonomous visual navigation.
\newblock {\em IEEE Robotics and Automation Letters}, 8(4):2054--2061, 2023.

\bibitem{chen2023}
Z.~Chen, D.~Pushp, J.M. Gregory, and L.~Liu.
\newblock Pseudo-trilateral adversarial training for domain adaptive traversability prediction.
\newblock {\em Autonomous Robots}, 2023.

\bibitem{CALI_2022_RSS}
Zheng Chen, Durgakant Pushp, and Lantao Liu.
\newblock {CALI: Coarse-to-Fine ALIgnments Based Unsupervised Domain Adaptation of Traversability Prediction for Deployable Autonomous Navigation}.
\newblock In {\em Proceedings of Robotics: Science and Systems}, 2022.

\bibitem{AB_Dame}
A.~Dame and E.~Marchand.
\newblock A new information theoretic approach for appearance-based navigation of non-holonomic vehicle.
\newblock In {\em 2011 IEEE International Conference on Robotics and Automation}, pages 2459--2464, 2011.

\bibitem{fox1997dynamic}
D.~Fox, W.~Burgard, and S.~Thrun.
\newblock The dynamic window approach to collision avoidance.
\newblock {\em IEEE Robotics \& Automation Magazine}, 4(1):23--33, 1997.

\bibitem{gupta2017unifying}
S.~Gupta, D.~Fouhey, S.~Levine, and J.~Malik.
\newblock Unifying map and landmark based representations for visual navigation.
\newblock {\em arXiv preprint arXiv:1712.08125}, 2017.

\bibitem{hirose2020probabilistic}
N.~Hirose, S.~Taguchi, F.~Xia, R.~Martin, Y.~Tahara, M.~Ishigaki, and S.~Savarese.
\newblock Probabilistic visual navigation with bidirectional image prediction.
\newblock {\em arXiv preprint arXiv:2003.09224}, 2020.

\bibitem{manderson2020vision}
T.~Manderson, J.~C.~G. Higuera, S.~Wapnick, J.~Tremblay, F.~Shkurti, D.~Meger, and G.~Dudek.
\newblock Vision-based goal-conditioned policies for underwater navigation in the presence of obstacles.
\newblock {\em arXiv preprint arXiv:2006.16235}, 2020.

\bibitem{Ort2018MNFramework}
T.~Ort, L.~Paull, and D.~Rus.
\newblock Autonomous vehicle navigation in rural environments without detailed prior maps.
\newblock In {\em 2018 IEEE International Conference on Robotics and Automation (ICRA)}, pages 2040--2047, 2018.

\bibitem{Pushp_2022}
D.~Pushp, S.~Kalhapure, K.~Das, and L.~Liu.
\newblock Uav-miniugv hybrid system for hidden area exploration and manipulation.
\newblock 2022.

\bibitem{shen2019situational}
W.~B. Shen, D.~Xu, Y.~Zhu, L.~J. Guibas, F.~Li, and S.~Savarese.
\newblock Situational fusion of visual representation for visual navigation.
\newblock In {\em IEEE/CVF International Conference on Computer Vision}, pages 2881--2890, 2019.

\bibitem{Srivastava_2021}
R.~Srivastava, R.~Lima, C.~Shinde, D.~Pushp, and K.~Das.
\newblock Estimation and control for autonomous uav system to neutralize unknown aerial maneuvering target.
\newblock In {\em 2021 Seventh Indian Control Conference (ICC)}, pages 117--122, 2021.

\bibitem{Strudel_2021_ICCV}
R.~Strudel, R.~Garcia, I.~Laptev, and C.~Schmid.
\newblock Segmenter: Transformer for semantic segmentation.
\newblock In {\em Proceedings of the IEEE/CVF International Conference on Computer Vision (ICCV)}, pages 7262--7272, October 2021.

\bibitem{SB_Zhang}
M.~Zhang, Y.~Li, and J.~Yang.
\newblock Autonomous visual navigation guided by path boundaries for mobile robot.
\newblock In {\em 2008 International Conference on Computer Science and Software Engineering}, volume~6, pages 344--348, 2008.

\end{thebibliography}
}
\end{document}